\documentclass[a4paper]{svproc}
\usepackage{makeidx}         % allows index generation
\usepackage{graphicx}        % standard LaTeX graphics tool

\usepackage{multicol}        % used for the two-column index
\usepackage[bottom]{footmisc}% places footnotes at page bottom

\usepackage{subcaption}
\usepackage{newtxtext}       % 
\usepackage[varvw]{newtxmath}       % selects Times Roman as basic font
\usepackage{xspace}
\usepackage{multirow}
\usepackage{wrapfig}
\usepackage{enumitem}

\usepackage[normalem]{ulem} % [normalem] removes the underlines in the reference section.
\usepackage{color}
\usepackage[per-mode=symbol]{siunitx}
\usepackage{cite}
\usepackage{array}
\usepackage{url}

\usepackage[colorlinks,bookmarksopen,bookmarksnumbered]{hyperref}
\usepackage{xcolor}
\usepackage[table]{xcolor} 
\usepackage{colortbl}  
\definecolor{lightblue}{rgb}{0,0.2,1}
\hypersetup{
    colorlinks=true,%
    citebordercolor=white,%
    filebordercolor=white,%
    linkbordercolor=white,%
    urlbordercolor=white,%
    citecolor=lightblue,%
    linkcolor=lightblue,%
    urlcolor=lightblue%
}

\usepackage[normalem]{ulem} % [normalem] removes the underlines in the reference section. % for \sout

\newcolumntype{C}[1]{>{\centering\arraybackslash}p{#1}}

\def\subparagraph{} % because IEEE classes don't define this, but titlesec assumes it's present
\usepackage[compact]{titlesec} 
\titlespacing*{\section}{0pt}{2pt}{2pt} % {left}{before}{after}
\titlespacing*{\subsection}{0pt}{2pt}{2pt} % {left}{before}{after}
\titlespacing*{\subsubsection}{8pt}{1pt}{2pt} % Adjust {left}{before}{after} as needed
\usepackage[font=small, skip=2pt]{caption}
\usepackage[skip=1pt]{subcaption}    % Reduce vertical gap between subfigures and captions

\usepackage[nodisplayskipstretch]{setspace}
\usepackage{eso-pic}

\AddToShipoutPictureBG*{%
  \AtPageUpperLeft{%
    \hspace{0.5\paperwidth}%
    \raisebox{-1.5cm}{%
      \makebox[0pt][c]{%
        \footnotesize Accepted for publication in the 19th International Symposium on Experimental Robotics (ISER), 2025%
      }%
    }%
  }%
}

\begin{document}

\mainmatter%%%%%%%%%%%%%%%%%%%%%%%%%%%%%%%%%%%%%%%%%%%%%%%%%%%%%%%
\title{From Zero to High-Speed Racing: An Autonomous Racing Stack 
\vspace{-10pt}}
% \titlerunning{Autonomous Racing Stack}
% From Zero to High-Speed Racing: Design and Implementation of an Autonomous Racing Stack

\author{Hassan Jardali \and Durgakant Pushp \and Youwei Yu \and Mahmoud Ali \and Ihab S. Mohamed \and Alejandro Murillo-Gonzalez \and Paul D. Coen \and Md. Al-Masrur Khan \and Reddy Charan Pulivendula \and Saeoul Park \and Lingchuan Zhou \and Lantao Liu
\vspace{-5pt}
}

\authorrunning{H. Jardali et al.}
% (feature abused for this document to repeat the title also on left hand pages)}

\institute{Luddy School of Informatics, Computing, and
Engineering at Indiana University, Bloomington.}

\maketitle

\vspace{-15pt}
\begin{abstract}
    High-speed, head-to-head autonomous racing presents substantial technical and logistical challenges, including precise localization, rapid perception, dynamic planning, and real-time control—compounded by limited track access and costly hardware. This paper introduces the Autonomous Race Stack (\texttt{ARS}), developed by the IU Luddy Autonomous Racing team for the Indy Autonomous Challenge (IAC). We present three iterations of our \texttt{ARS}, each validated on different tracks and achieving speeds up to \SI{260}{\kilo\meter\per\hour}. Our contributions include: (i) the modular architecture and evolution of the \texttt{ARS} across \texttt{ARS1}, \texttt{ARS2}, and \texttt{ARS3}; (ii) a detailed performance evaluation that contrasts control, perception, and estimation across oval and road-course environments; and (iii) the release of a high-speed, multi-sensor dataset collected from oval and road-course tracks. Our findings highlight the unique challenges and insights from real-world high-speed full-scale autonomous racing.
\end{abstract}
\vspace{-12pt}

% -------------------------------------------------------------------------------------------------------
\section{Introduction}
\label{sec:intro}
Developing large-scale autonomous systems, particularly those highlighted by competitions such as the DARPA Challenges, faces considerable technical and logistical challenges. High-speed autonomous racing exemplifies these difficulties, as it demands precise localization, rapid perception, and real-time planning and control under extreme dynamic conditions. However, significant hardware investment and limited track time restrict development, making well-planned real-world tests crucial. 
This paper describes the development of the Autonomous Race Stack (\texttt{ARS}) by our team for participation in the IAC, presenting its iterative evolution, the rationale behind key design choices, % HJ: I commented this cause we didn't mention clearly the reasons behind choosing a method over  ?
and performance insights from real-world testing on both oval and road-course tracks.
\begin{wrapfigure}{r}{0.4\textwidth}
    \vspace{-10pt}
    \centering
    \includegraphics[width=\linewidth, trim=200 300 0 500, clip]{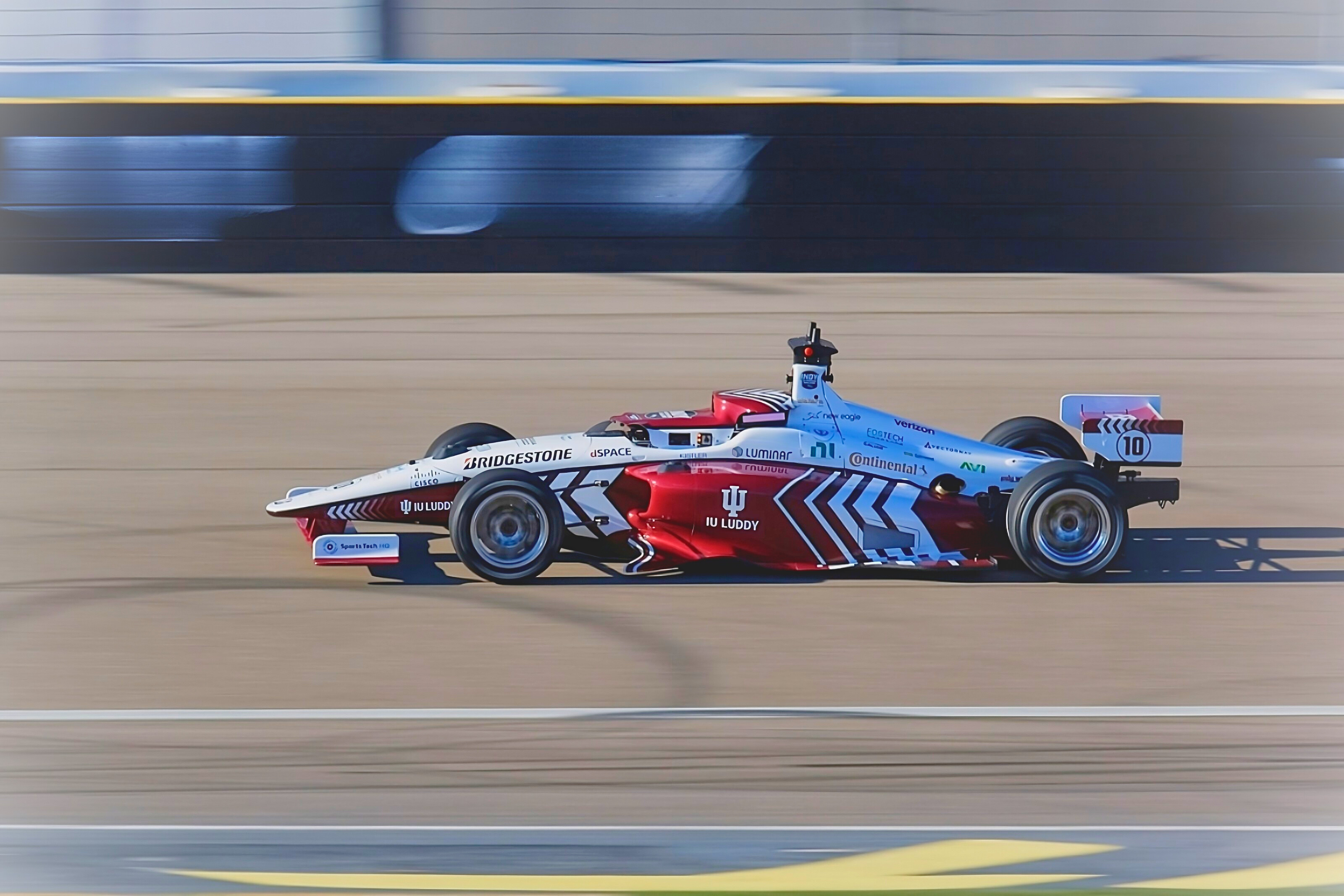} 
    \caption{\small IU-Luddy's IAC AV24 racing vehicle at LVMS.} 
    \label{fig:IU_Luddy_av24-label} \vspace{-10pt}
\end{wrapfigure}
More precisely, this paper details the development and evaluation of three successive \texttt{ARS} iterations: \texttt{ARS1}, validated at the Indianapolis Motor Speedway (IMS) at speeds up to \SI{205}{\kilo\meter\per\hour}; \texttt{ARS2}, demonstrated at the Las Vegas Motor Speedway (LVMS) reaching \SI{260}{\kilo\meter\per\hour}, and later tested on the Las Vegas Road Course (LVRC); and \texttt{ARS3}, designed for head-to-head autonomous racing on road-course tracks and partially validated at both LVMS and LVRC.
We discuss the evolution of the stack across these iterations, highlighting key architectural updates, technical challenges, and insights gained through real-world testing.

% \texttt{ARS1} incorporates a filter-based state estimation module that fuses satnav and inertial measurements and employs a pure-pursuit controller to follow the raceline generated by the minimum-curvature constraint~\cite{heilmeier2020minimum}. \texttt{ARS2}, built upon \texttt{ARS1}, includes modules needed for multi-vehicle racing, includes Radar-based perception, Adaptive Cruise Control (ACC) and Anti-lock Braking System (ABS), and it also utilizes a linear Model Predictive Controller (MPC) for lateral control. Finally, \texttt{ARS3} fuses exteroceptive sensors into optimization-based state estimation~\cite{iSAM2}, enhances opponent reasoning via Radar and LiDAR coupled perception. It also includes a coupled stochastic MPC for longitudinal and lateral control.
The initial version, \texttt{ARS1}, incorporates filter-based state estimation fusing Global Navigation Satellite System (GNSS) and inertial measurements, and employs a Pure Pursuit controller to track racelines generated via the minimum curvature method~\cite{heilmeier2020minimum}. Building on this foundation, \texttt{ARS2} introduces multi-vehicle capabilities, including radar-based perception, Adaptive Cruise Control (ACC), and an Anti-lock Braking System (ABS), while also utilizing a linear Model Predictive Controller (MPC) for lateral control. The latest iteration, \texttt{ARS3}, integrates exteroceptive sensor data into an optimization-based state estimation framework~\cite{iSAM2}, enhances opponent reasoning through coupled radar and LiDAR perception, and features an integrated stochastic MPC for combined longitudinal and lateral control.

% \ihab{\sout{Prior works featuring full-scale autonomous racing stacks~\cite{unimore_full, betz2023tum, jung2023autonomous} have demonstrated impressive results in previous seasons on the Dallara AV-21, the previous version on the IAC's official autonomous car;
% % \hassan{expand here more about each of the other work.} 
% This study distinguishes itself from previous literature by detailing the evolution of our software stack through successive iterations and by presenting a more in-depth analysis of autonomous racing on challenging road courses and, thereby expanding the focus beyond the oval-track-based research. Furthermore, we present preliminary results from our sampling-based controller tests and an analysis of vehicle spin incidents recorded at LVRC.} 
Numerous teams have participated in the IAC, each developing its own full-scale autonomous racing stack, with several prior works~\cite{unimore_full, betz2023tum, jung2023autonomous} demonstrating impressive performance on the Dallara AV21, the IAC’s previous-generation autonomous vehicle. 
This work distinguishes itself by documenting the systematic evolution of our software across three iterations, providing a detailed performance analysis on both oval and road-course tracks—thereby expanding the scope beyond predominantly oval-focused studies. 
Additionally, we analyze a spin incident recorded at LVRC.
% Additionally, we present preliminary results from our sampling-based controller and analyze a critical spin incident recorded at LVRC.
We build on these advancements to make the following contributions:
\begin{enumerate}[leftmargin=*] \vspace{-3pt}
    \item \textbf{Autonomous Race Stack:} We introduce a comprehensive \texttt{ARS} designed for solo and multi-vehicle racing on both oval and road-course tracks, validated on a full-scale IAC AV24 platform, as shown in Fig.~\ref{fig:IU_Luddy_av24-label}. The system has demonstrated stable performance at speeds reaching up to \SI{260}{\kilo\meter\per\hour}. 
    \item \textbf{Evaluation:} We present a performance evaluation of each \texttt{ARS} subsystem, supplemented by analytical justifications and insights into our system updates.
    % that distinguish our work from existing studies in full-scale autonomous racing literature.
    \item \textbf{Dataset:} We provide high-speed sensor data collected from the IMS, LVMS, and LVRC tracks, highlighting the significant challenges of synchronizing and fusing information from cameras, radars, and LiDARs to achieve reliable localization, dynamics estimation, and opponent detection\footnote{See \href{https://iu-vail.github.io/IU-ARS/}{https://iu-vail.github.io/IU-ARS} for access to the open-source data and dashboard code.}.
\end{enumerate}

% Distinct from other controllers in the autonomous racing literature, we present a software stack that integrates a GPU-accelerated MPPI controller to enhance real-time performance. The stack also incorporates advanced safety features, including the Multi-Mode Control Manager (MCM), which evaluates the outputs of three controllers and dynamically switches control modes when necessary. (2) We detail our development and development  we have achieved a top-speed of 260 KM/H in a short testing time (x hours). We detail our rapid deployment approach for cost efficient system development. Investigation of state-of-the-art segmentation methods on the racing car. We discuss the reasons these methods proved ineffective and provide insights from field experiments to benefit future research.
\section{System Architecture}
% \section{System Architecture of the IU-Luddy Autonomous Racing Vehicle} % HJ: I don't prefer it here
\label{sec:technical_approach}

In this section, we introduce the IAC AV24 vehicle and then our software architecture. 

% ------------------------------------------
% \noindent\bfit{Hardware Architecture.}
\subsection{Hardware Architecture}
The IAC AV24 race car is based on the Dallara IL15 design and equipped with a \SI{2}{\liter} single-turbo inline-4 engine with \num{488} HP. Its cockpit is replaced by autonomous components, including a dSPACE Autera AutoBox with an Intel Xeon 3GHz 12-core CPU and an NVIDIA A5000 GPU. It runs Ubuntu 22.04 with ROS2 Humble and Cyclone DDS as a middleware for real-time communication and core development framework. The equipped sensors related to critical autonomy modules are as following: %, with details in the link provided:
\begin{itemize}
    \item[$\circ$] Three Luminar Iris LiDARs operating at \SI{10}{\hertz};
    \item[$\circ$] Six Mako G-319C cameras with a resolution of $2064 \times 1544$ at \SI{10}{\hertz};
    \item[$\circ$] Two Continental ARS548 4D Radars at \SI{20}{\hertz};
    \item[$\circ$] Two NovAtel PwrPak7D GNSS units (Top and Side) at \SI{20}{\hertz} with IMUs at \SI{125}{\hertz};
    \item[$\circ$] One VectorNav-310 GNSS unit at \SI{5}{\hertz} with IMU at \SI{200}{\hertz}.
\end{itemize}
This diverse sensor configuration enables robust state estimation and perception, supporting the vehicle's operation across both oval and road-course racing environments.

\begin{figure}[t!] 
    % \vspace{-5pt}
    \centering
    \includegraphics[width=\linewidth, trim=0 0 0 0, clip]{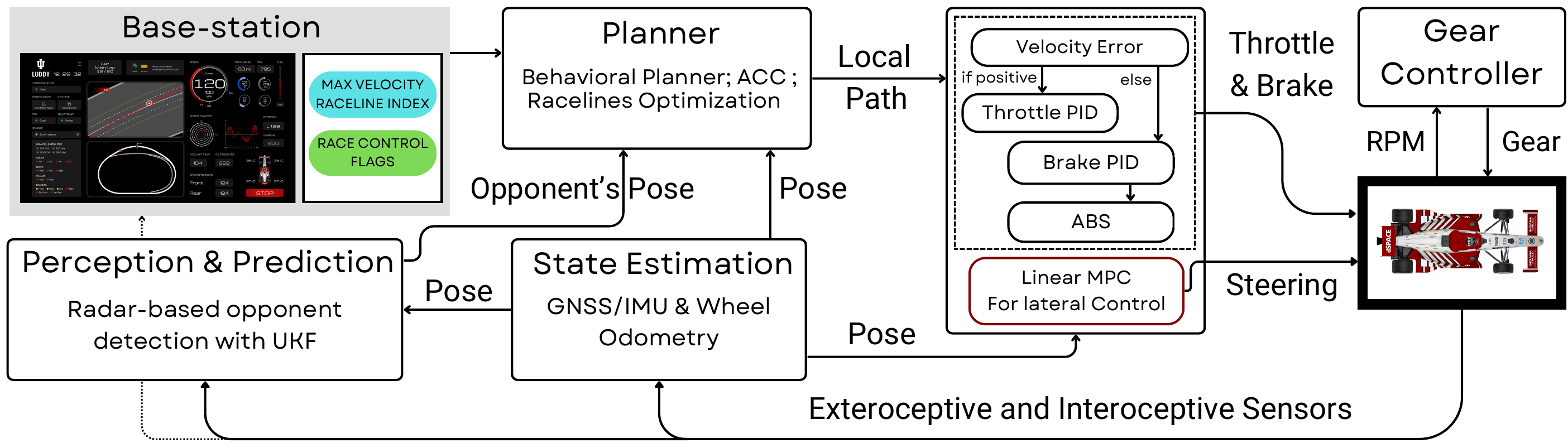}
    \vspace{-0.2in}
    % \caption{\small System architecture of \texttt{ARS2} including state estimation, perception, planning, and control modules.}
    \caption{\small System architecture of \texttt{ARS2} with core modules' details.}
    \label{fig:system-diagram}
    \vspace{-0.15in}
\end{figure}

\subsection{Autonomy Stack Overview} 
Building on the hardware capabilities of the IAC AV24 autonomous race car, our autonomy stack integrates multiple high-performance modules to enable safe and competitive racing.
% across diverse track configurations. 
The vehicle operates on both oval and road-course tracks under two primary race formats: a solo-vehicle (SV) format, defined as a time trial where a single car records its fastest lap time, and a head-to-head (H2H) format that includes two cars on non-intersecting lines.
The autonomy stack is composed of several interconnected modules. A state estimation module fuses measurements from multiple onboard sensors to estimate the ego vehicle's pose and velocity. This is complemented by a perception and prediction system responsible for detecting nearby opponents and forecasting their future trajectories.
Based on this situational awareness, a behavioral planner determines whether to maintain a distance with the leading car or initiate an overtaking maneuver. A path planner then generates dynamically feasible trajectories, which are followed by a control module that issues low-level actuation commands: steering for lateral control, and throttle, brake, and gear shifting for longitudinal control.
% A remote base-station interface allows the operator to monitor system status and performance in real-time, as well as issue race-time parameters such as maximum target speed. A dashboard for real-time data visualization was implemented and updated subsequently with the stack updates. The overall \texttt{ARS2} system architecture is illustrated in Fig.~\ref{fig:system-diagram}.
A remote base-station interface enables real-time monitoring of system status and performance, while also allowing the operator to issue race-time parameters such as maximum target speed. The dashboard for live data visualization was continuously updated in line with stack revisions. The complete \texttt{ARS2} system architecture is shown in Fig.~\ref{fig:system-diagram}.

\begin{table}[htbp]
\vspace{-10pt}
\centering
{\small
\caption{Key architectural differences between \texttt{ARS1}, \texttt{ARS2}, and \texttt{ARS3}.}
\label{tab:ars_summary}
\rowcolors{1}{gray!25}{white}
\begin{tabular}{|l|c|c|c|}
\hline
\rowcolor[HTML]{CBCBCB}
\textbf{Feature} & \texttt{ARS1} & \texttt{ARS2} & \texttt{ARS3} \\ \hline
\textbf{Racing Scenario} & SV & SV \& 
 H2H & SV \& H2H \\ \hline
\textbf{Track} & IMS & LVMS \& LVRC & LVMS \& LVRC \\ \hline
% \textbf{Track Hours} & 11 hours & 6.5+13.5 hours & Approx.2 hour \\ \hline
% \textbf{Kilometers Traveled} & 325 km & 166 km-TBD & TBD \\ \hline
\textbf{Top Speed} & \SI{205}{\kilo\meter\per\hour} & \SI{260}{\kilo\meter\per\hour} (LVMS) \SI{195}{\kilo\meter\per\hour} (LVRC) & \SI{72}{\kilo\meter\per\hour} (both) \\ \hline
\textbf{Estimation} & GPS/INS & \texttt{ARS1} \& Wheel Odometry & \texttt{ARS2} \& LiDAR \\ \hline
\textbf{Perception} & N/A & Radar &  Radar \& LiDAR  \\ \hline
% \textbf{Prediction} & N/A & UKF & MPC-based \\ \hline
\textbf{Lateral Controller} & PP & MPC & MPPI \\ \hline
% \textbf{Validation Status} & Validated & Validated & To be validated \\ \hline
\end{tabular} 
% \vspace{-12pt}
}
\end{table}

%---------------------------
\section{Evolution of the IU-Luddy Autonomous Racing Stack}
In this section, we describe the evolution of our software stack across its three iterations by detailing the key updates within each module. Additionally, the differences between iterations are highlighted in Table~\ref{tab:ars_summary}.

% To handle the increasing complexity and performance demands of autonomous racing, our team iteratively developed three versions of the Autonomous Race Stack (ARS). Each version incorporates lessons learned from prior implementations and introduces new capabilities tailored to different race formats and track types. We describe each ARS version below, highlighting major architectural upgrades, experimental results, and deployment challenges.
% ----------------------------------------------------
% \subsection{\texttt{ARS1}: Single-Vehicle Time Trials} %  at IMS
\subsection{\texttt{ARS1}: Time Trials} %  at IMS
\texttt{ARS1}, the initial iteration of our autonomous racing stack focused on single-vehicle racing, was validated at IMS in September 2024. The primary design goal was to create a system that is both minimally complex and inherently stable. The \textit{state estimation} module employs an error-state extended Kalman filter (ESKF), fusing NovAtel-Top GNSS with both NovAtels' inertial data via Manifold~\cite{yu23mimu}. The ESKF propagates the state $x \in \mathbb{R}^{18}$, including position $(\in \mathbb{R}^{3})$, velocity $(\in \mathbb{R}^{3})$, orientation $(\in \mathbb{R}^{3})$, and their first-order derivatives.
The \textit{planning module} generates a global raceline offline using the minimum curvature method~\cite{heilmeier2020minimum}, while the online planner outputs the nearest path segment in the car's frame. 
% \textit{Control} is achieved via a Pure Pursuit (PP) steering controller \cite{coulter1992implementation}, which calculates the steering angle $\delta$ using the geometric relation: $\delta = \arctan\left( \frac{2L \sin \alpha}{L_{fw}} \right)$, where $L$ is the vehicle wheelbase, $\alpha$ is the angle to the target point on the path, and $L_{fw}$ is the adaptive look-ahead distance determined by the current velocity $v_{car}$; this is coupled with PID-based longitudinal control where the velocity error, $e_v = v_{target} - v_{car}$ (with $v_{target} = \max(v_{max}, v_{raceline})$), governs PID activation: throttle PID if $e_v > 0$, and brake PID if $e_v < - \delta_{db} \, v_{target}$ (where $\delta_{db}$ is a deadband coefficient). Gear shifting is regulated based on the engine's RPM. 
\textit{Control} is achieved via a Pure Pursuit (PP) steering controller
% ~\cite{coulter1992implementation}
, which computes the steering angle $\delta$ using the geometric relation:
$
\delta = \arctan\left( {2L \sin \alpha}/{L_{fw}} \right),
$
where $L$ is the vehicle wheelbase, $\alpha$ is the heading angle to the target point on the path, and $L_{fw}$ is the adaptive look-ahead distance determined by the current vehicle velocity $v_{\text{car}}$.
This is coupled with a PID-based longitudinal controller, where the velocity error $e_v = v_{\text{target}} - v_{\text{car}}$, with $v_{\text{target}} = \max(v_{\text{max}}, v_{\text{raceline}})$, governs PID activation: the throttle PID is used if $e_v > 0$, and the brake PID is triggered if $e_v < -\kappa_{\text{db}} \; v_{\text{target}}$, where $\kappa_{\text{db}}$ is a deadband coefficient. Gear shifting is regulated based on engine RPM.
% -----------------------------------------------------------

% \subsection{\texttt{ARS2}: Toward Multi-Agent Racing with Advanced Planning and Control} % at LVMS and LVRC
\subsection{\texttt{ARS2}: Toward Multi-Agent Racing via Advanced Planning and Control}
\label{sec:ars2}
\texttt{ARS2} extends the capabilities of \texttt{ARS1}, introducing enhancements to the state estimation and control modules and incorporating new components essential for multi-vehicle racing. These additions include a perception module for opponent detection, a prediction module, a behavioral planner, and adaptive cruise control. This version was validated on LVMS and LVRC in January and April 2025, respectively.

The \textit{state estimation} module in \texttt{ARS2} augments the existing ESKF with vehicle dynamics odometry. To preserve the accuracy of the yaw angle estimation, roll and pitch angles are estimated independently. The vehicle single-track model incorporates wheel speed and steering angle measurements to refine the state's velocity. We smooth the inconsistency caused by possibly long-term wheel odometry dead-reckoning via an inverse multi-quadratic (IMQ) weighting function~\cite{pmlr-v235-duran-martin24a}, denoted as $ w(x_t, \hat{x}_t) = \left( 1 + {\| x_t - \hat{x}_t \|_2^2}/{c^2} \right)^{-1/2} $, with the soft threshold $ c $, the predicted robot state $ \hat{x} $, and the measurement $ x $. IMQ is applied to the Kalman gain $ K $, thereby influencing the update of state covariance $ \Sigma $:
\vspace{-1.8ex}
\begin{align*}\label{eqn:kalman-gain}
\Sigma_t^{-1} \leftarrow \Sigma_{t|t-1}^{-1} + w_t^2 H_t^\top R_t^{-1} H_t, \quad K_t \leftarrow w_t^2 \Sigma_t H_t^\top R_t^{-1},
\end{align*}
\vspace{-0.28in}

\noindent with measurement matrix $ H_t $ and its noise covariance $ R_t $. Additionally, we create a safety state machine so that the supplementary NovAtel-Side (i.e., with antennas on two sides) will take over if the main NovAtel-Top loses signals for \SI{0.5}{\second}.

\begin{wrapfigure}{r}{0.42\textwidth} % 'r' for right, 'l' for left
    \vspace{-5pt}
    \centering
    \includegraphics[scale=0.7]{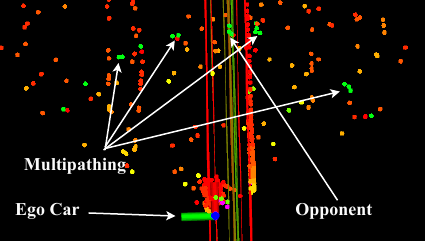}
    % \vspace{-8pt}
    \caption{\small Front Radar Pointcloud.} 
    % in a H2H trial at LVMS.}
    % \vspace{-4pt}
    \label{fig:radar_fig} 
    \vspace{-13pt}
\end{wrapfigure}

For \textit{perception}, \texttt{ARS2} employs a radar-based \textit{opponent detection} algorithm. The choice of radar over LiDAR for this task stems from its capacity for direct velocity measurement, its higher update rate, and its lower computational load.
% The raw radar pointcloud is preprocessed to reduce the noise,
% then the static objects are filtered out, while the moving objects are clustered based on 3D spatial position and radial velocity. Track boundaries are used to filter out clusters outside the track caused by radar multipathing. 
The radar outputs a 4D point cloud \( P = \{p_i\}_{i=1}^N \), where each point \( p_i \) is defined by its Cartesian coordinates and radial velocity, along with multiple signal quality metrics. An initial filtering step is applied using predefined thresholds for these metrics, including signal-to-noise ratio (SNR), existence probability, peak detection confidence, received signal strength (RSS), and ambiguity identification.
Subsequently, static objects—identified by radial velocities matching the negative of the ego-car velocity—are filtered out. 
% While, the remaining moving objects are clustered based on their 3D spatial coordinated and radial velocity using the \textsc{DBSCAN} algorithm~\cite{ester1996density}.
The remaining dynamic objects are then clustered based on their 3D spatial coordinates and radial velocity using the \textsc{DBSCAN} algorithm~\cite{ester1996density}.
A valid cluster \( C \) is retained only if it contains at least \( N_{\text{min}} \) points, increasing robustness against noise.
Clusters undergo a {coordinate transformation} from the ego-vehicle's radar frame to the global frame, then filtered using the {track boundary} as follow $p_i \in \Omega_{\text{outer}} \setminus \Omega_{\text{inner}}$, where $\Omega_{\text{outer}}$ and $\Omega_{\text{inner}}$ denote the outer and inner track boundaries set, respectively. This step efficiently filters out clusters outside the track caused by radar multipathing. 
Fig.~\ref{fig:radar_fig} shows a radar observation, illustrating both the opponent vehicle and multipath reflections, with point color representing radial velocity.
% Following this filtering stage, historical detection data are utilized to predict the opponent's position and velocity, and the cluster most consistent with this prediction is selected. 
A cost function  
% \( J \) 
is computed between the current opponenet detections at time-step \( k \) and the prior estimate at \( k-1 \), incorporating both the opponent's 3D pose 
% \( \mathbf{x}_{op} \)
and velocity
% \( v_{op} \); $    J = \alpha \| \mathbf{x}_{op,k} - \mathbf{x}_{op,k-1} \| + \beta | v_{op,k} - v_{op,k-1} |$
. The detection with the minimal cost is then propagated to an Unscented Kalman Filter (UKF)~\cite{wan2000unscented}, that employs
a point-mass motion model, for robust \textit{opponent tracking}. 
% \textit{Opponent tracking} is achieved using UKF with point-mass motion model that integrate new detection with prior estimates of the opponent's state. 
This approach ensures reliable opponent tracking, even in scenarios where detections are temporarily unavailable. 
In multi-vehicle scenarios within \texttt{ARS2}, participating vehicles typically operate on parallel racelines to ensure safety. The \textit{behavioral planning} module processes opponent detection data from the perception module to make high-level strategic decisions. It selects between two primary operational modes: \textit{Follow mode} or \textit{Attack mode}, with the choice contingent upon inputs from race control and adherence to competition regulations.
In the Follow mode, an ACC system is employed. The ACC dynamically modulates the ego vehicle's longitudinal speed to maintain a predefined, safe separation distance from the leading vehicle. Conversely, activation of the Attack mode authorizes the ego vehicle to accelerate, thereby initiating an overtaking maneuver.
The ACC system calculates a preliminary desired longitudinal velocity, $v'_{des}$, for the ego vehicle. This calculation considers the state of both the ego and leading vehicles, and is given by:
$v'_{des} = v_{l} + K_{d} \left( d - \left(d_{0} + \max(0, \tau_{des} (v_{e} - v_{l}))\right) \right)$
where $v'_{des}$ is the preliminary desired longitudinal velocity, $v_{l}$ is the leading vehicle's velocity, $v_{e}$ is the ego vehicle's velocity, $d$ is the longitudinal separation, $d_{0}$ is the target separation distance, $\tau_{des}$ is the desired inter-vehicle time gap, and $K_{d}$ is a gain parameter. 
% The commanded velocity adjustment is then limited by the vehicle's acceleration/deceleration capabilities and must remain non-negative.

For \textit{longitudinal control}, an ABS was implemented for road-course racing; it operates by continuously calculating the slip ratio for each wheel—defined as the normalized difference between the vehicle's longitudinal speed and the wheel's tangential speed, $\lambda = (V_{veh} - \omega R) / V_{veh}$ and then employs a PID controller to regulate brake pressure and track the optimal slip ratio, thereby preventing wheel lock-up and maximizing deceleration. By avoiding lock-up during aggressive deceleration, particularly before turns, ABS maintains optimal tire traction. 
% This maximizes braking effectiveness, preserves steerability to mitigate understeer, and enhances vehicle stability, thereby reducing the risk of spinning.    
This enhances braking effectiveness, preserves steerability to mitigate understeer, and improves vehicle stability, ultimately reducing the risk of spinning.
% On the \textit{lateral control} level
With respect to \textit{lateral control}, \texttt{ARS2} employs a linear Model Predictive Control (LMPC) framework. The controller is formulated based on the lateral single-track dynamic model described in \cite{rajamani2011vehicle} and explicitly accounts for road banking angles, which can reach up to \SI{20}{\degree} at LVMS. The LMPC is implemented using the ACADOS optimization package \cite{verschueren2022acados}.
% and runs at \SI{50}{\hertz}. 
% -----------------------------------------------------------
% \subsection{\texttt{ARS3}: Lidar-based odometry and perception and stochastic control}
\subsection{\texttt{ARS3}: LiDAR-Based Odometry, Perception, and Stochastic Control}
\texttt{ARS3} represents the latest proposed iteration of our autonomous racing software stack, featuring significant advancements in key areas such as an enhanced LiDAR-based state estimation and perception modules, alongside the implementation of a stochastic sampling-based controller, specifically a Model Predictive Path Integral (MPPI) controller \cite{williams2017model}. While \texttt{ARS3} has been deployed and subjected to initial testing at LVRC, full integration and validation of all updated modules were constrained by limited track access and debugging windows. 
% This section will provide an overview of the \texttt{ARS3} architecture.

In \texttt{ARS3}, we aim to improve the \textit{state estimation} accuracy by tightly coupling the rich sensor suite through nonlinear incremental optimization, specifically iSAM2~\cite{iSAM2}, where its chordal Bayes tree structure can leverage diverse sensory inputs. To improve the robustness of simultaneous localization and mapping (SLAM), we leverage RTK to construct pointcloud maps offline, which are subsequently used as priors for LiDAR registration~\cite{vizzo2023ral}. Three LiDAR sensors operate asynchronously to enhance state estimation, particularly in orientation. Moreover, they serve as the primary source of state estimation when all GNSS devices become unavailable.
% For the \textit{perception} module, \texttt{ARS3} utilizes LiDAR to address key limitations of radar, improving the detection of static opponents on the track (such as vehicles stopped by safety checks).
For the \textit{perception} module, \texttt{ARS3} leverages LiDAR to overcome key limitations of radar, enhancing the detection of static obstacles on the track, such as vehicles halted due to safety checks.
% \sout{For \textit{control}, the architecture features a GPU-accelerated MPPI controller as the primary mechanism for generating longitudinal acceleration and steering commands \cite{mohamed2022autonomous}. It was initially tested with a kinematic model, and later updated to a single-track dynamic model. A PID controller then translates the longitudinal commands into throttle and brake signals.}
For \textit{control}, the architecture features a GPU-accelerated MPPI controller as the primary mechanism for generating longitudinal acceleration and steering commands \cite{mohamed2022autonomous}. 
It was initially tested with the kinematic bicycle model, where a PID controller was used to convert longitudinal acceleration into throttle and brake signals. In the updated version, the controller employs a single-track dynamic model that directly outputs throttle, brake and throttle commands.

% while concurrently running the two previous lateral controller serve as crucial backups due to potential GPU failures from high-RPM vibrations or if MPC fails.
% \hassan{What about prediction, we don't need it for now.} 
% \textit{Opponent behavioral prediction} is achieved using a model predictive controller as the surrogate opponent, whose behavior is modeled online via historical states rather than relying on constant velocity or non-parametric Gaussian Process estimators.

% a local path planner based on \textit{Frenet-Frame} \cite{werling2010optimal} creates dynamically feasible paths for switching between racelines.
% The primary control algorithm is an MPPI approach running on the GPU, initially tested with a kinematic model and now updated to incorporate a single-track dynamic model. 
% This structure also enables the exploration of advanced variants, such as U-MPPI~\cite{mohamed2025towards}, to address uncertainties in dynamics modeling and opponent overtaking planning. 

\section{Experiments}
\label{sec:experiments} 

% We highlight the experiments done at IMS in September 2024 (\texttt{ARS1}) and LVMS in January 2025 and at LVRC in April 2025 using (\texttt{ARS2}). (\texttt{ARS3}) preleminary results are shown in the last part of this section.  
% \ihab{\sout{This paper highlights experiments conducted with \texttt{ARS1} at the IMS in September 2024, as well as those utilizing \texttt{ARS2} at LVMS in January 2025 and at LVRC in April 2025. Preliminary results for \texttt{ARS3} are presented in the concluding part of this section.}
This section presents a series of experiments conducted to evaluate the successive iterations of our autonomous racing software stack. We report results from deployments of \texttt{ARS1} at the IMS in September 2024, \texttt{ARS2} at LVMS in January 2025 and LVRC in April 2025. Preliminary results for \texttt{ARS3} are also included at the end of this section.

% using React.js, Redux, Tailwind CSS, and Node.js, with D3.js for visualization. 
% Data is transmitted from the car via UDP, with Google Protocol Buffers handling serialization. 

% \noindent \bfit{State Estimation.} 
\subsection{State Estimation} 
\begin{figure}[t!]
\centering
\begin{subfigure}[b]{1\textwidth}
    \centering
    \includegraphics[width=\linewidth, trim=200 200 200 150, clip]{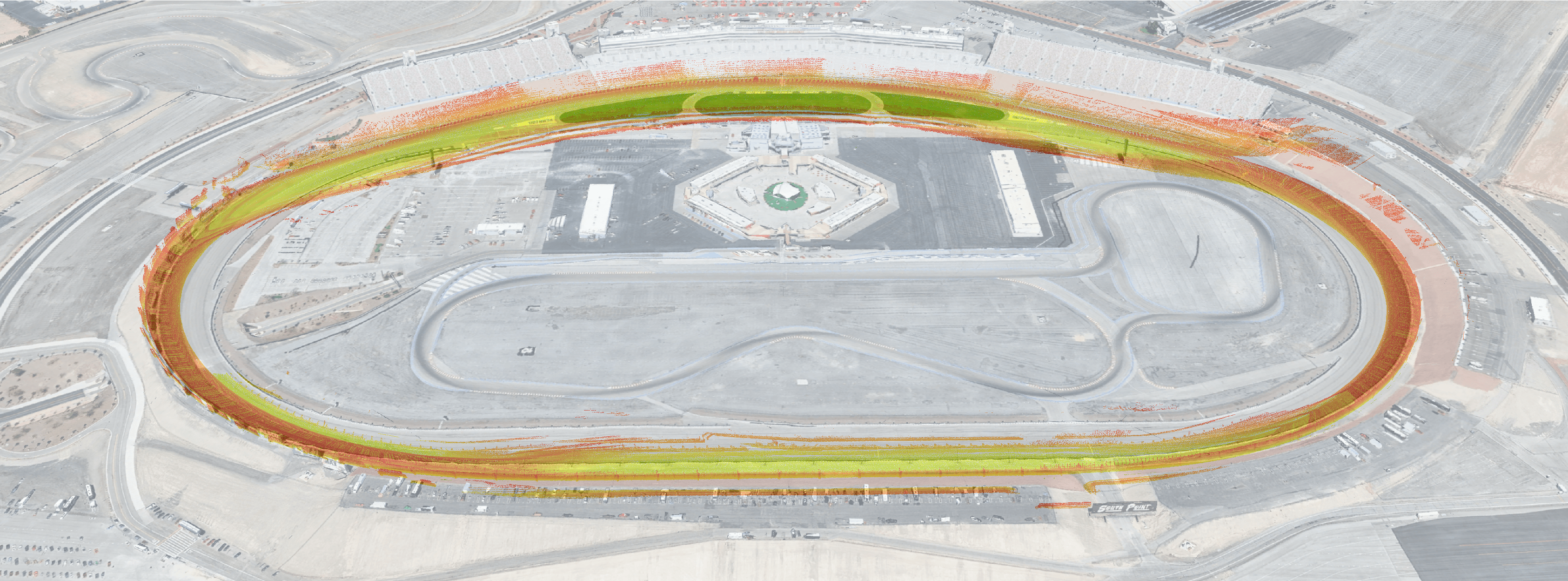} 
    \label{fig:est_gt_pos_lvrc}
    \vspace{-0.12in}
\end{subfigure}
\hfill
\begin{subfigure}[b]{1\linewidth}
    \centering
    \includegraphics[width=\linewidth, trim=0 164 0 166, clip]{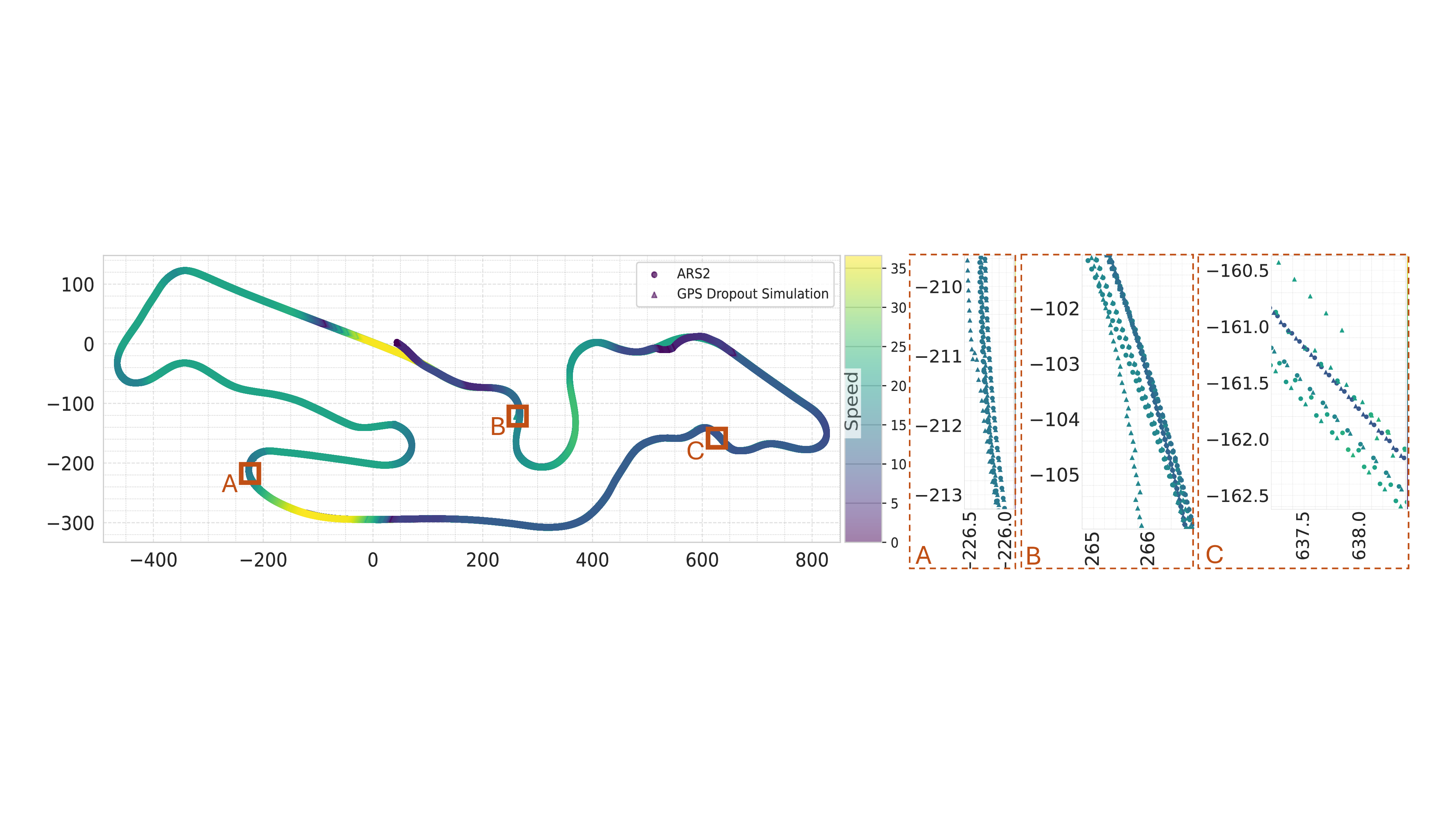} 
    \label{fig:est_gt_pos_lvrc}
\end{subfigure}
% 0.605 0.395
\vspace{-0.32in}
\caption{\small (Top) Pointcloud mapping at LVMS with poses estimated by dynamics-satnav-inertial fusion. (Bottom) GPS dropout simulations on track with zoomed-in view.}
\label{fig:lvms-pcl-dbd}
\vspace{-0.15in}
\end{figure}

We evaluate the localization performance using running statistics. Rather than relying on RTK as ground truth, we assess the satellite navigation (satnav)-inertial fusion during the \texttt{ARS1} deployment and the dynamics-satnav-inertial fusion during \texttt{ARS2}. Our metrics include heading and position jerks, defined as the third derivative of the heading and position over time. During \texttt{ARS1}, our satnav-inertial system achieves a heading jerk of \SI{407.6}{\degree\per\second\cubed} and a position jerk of \SI{1387.17}{\meter\per\second\cubed}. In \texttt{ARS2}, our dynamics-satnav-inertial fusion improves these to \SI{305.56}{\degree\per\second\cubed} and \SI{774.16}{\meter\per\second\cubed}, benefiting from the vehicle dynamics odometry based on the single-track model for enhanced consistency and robustness. 
% Since we used the same controller for localization tests, these motion stability metrics indicate the improvement in localization. 
Since the same controller was used for all localization tests, these motion stability metrics directly reflect the improvements in localization. 
Moreover, as shown in Fig.~\ref{fig:lvms-pcl-dbd}, the pointcloud overlapped with the LVMS map validates the scale consistency. Note that the pointcloud is projected solely based on dynamics-satnav-inertial fusion without any LiDAR odometry.

% During testing at LVRC, we experienced no RTK dropouts (greater than \SI{0.5}{\second}). 
During testing at LVRC, we experienced no RTK dropouts exceeding \SI{0.5}{\second}.
%Nonetheless, we intentionally disabled the GPS (i.e., no sensor outputs) to assess our localization robustness.
Nonetheless, to assess the robustness of our localization system, we intentionally disabled the GPS, resulting in no sensor outputs.
% Fig.~\ref{fig:lvms-pcl-dbd} showed a maximum jump of around \SI{0.5}{\meter} after disabling GPS for \num{6} seconds on both straight and turns, and the jump back to normal pose was smoothed by IMQ. 
As shown in Fig.~\ref{fig:lvms-pcl-dbd}, the system exhibited a maximum deviation of approximately \SI{0.5}{\meter} after \num{6} seconds of GPS denial on both straight segments and turns. The recovery to nominal pose was smoothly managed by the IMQ mechanism.
In terms of IMQ parameters, we use a vehicle speed-based soft threshold $ c $. Position uses $ \max(0.75, \min(3, v\cdot \Delta t)) $ and orientation uses $ \max(\frac{\pi}{36}, \min(\frac{\pi}{18}, \omega \cdot \Delta t)) $ with linear speed $ v $~[\si{\meter\per\second}] and angular velocity $ \omega $~[\si{\radian\per\second}].

\captionsetup[subfloat]{labelformat=empty}

% \noindent \bfit{Perception and Prediction.}
\subsection{Perception and Prediction}

\begin{figure}[h!]
    \centering
      \vspace{-5pt}
    % First subfigure
    \begin{subfigure}[b]{0.48\textwidth}
        \centering
        \includegraphics[scale=0.6]{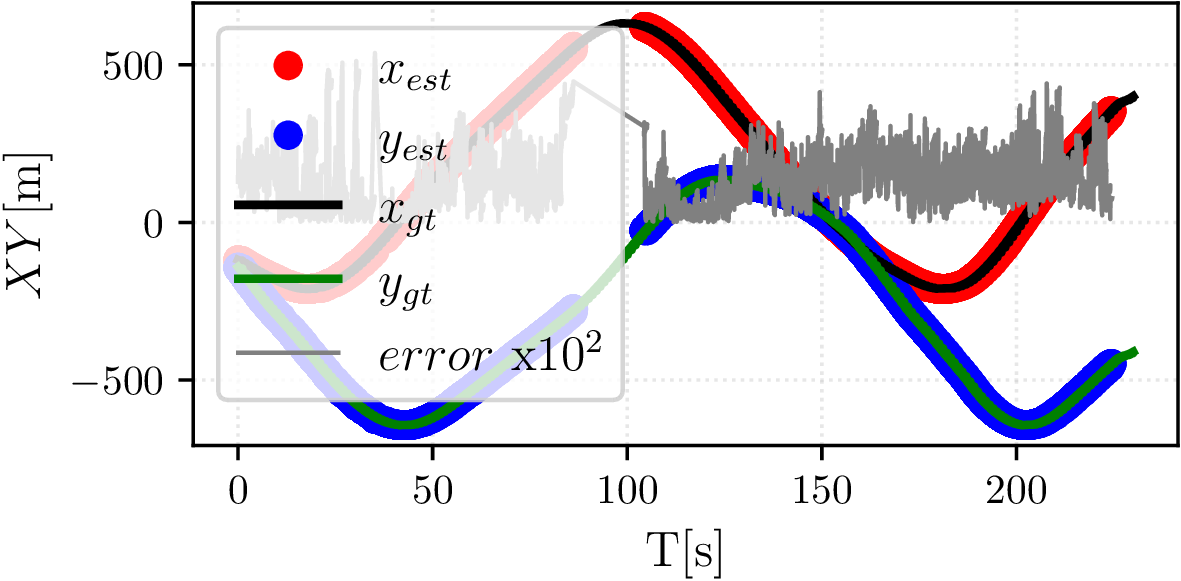} 
        % \caption{Estimated Vs ground-truth Position at LVMS}
        \caption{(a) LVMS}
        \label{fig:est_gt_pos_lvms}
    \end{subfigure}
    \hfill 
    % Second subfigure
    \begin{subfigure}[b]{0.48\textwidth} 
        \centering
        \includegraphics[scale=0.6]{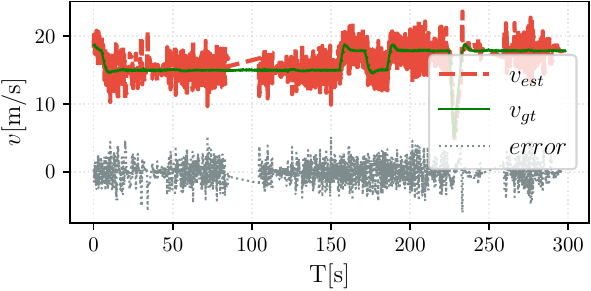}
        % \caption{Estimated Vs ground-truth Velocity at LVMS}
        \caption{(b) LVMS}
        \label{fig:est_gt_vel_lvms}
    \end{subfigure}
    \vfill
    % \vspace{5pt}
    \begin{subfigure}[b]{0.48\textwidth}
        \centering
        \includegraphics[scale=0.6]{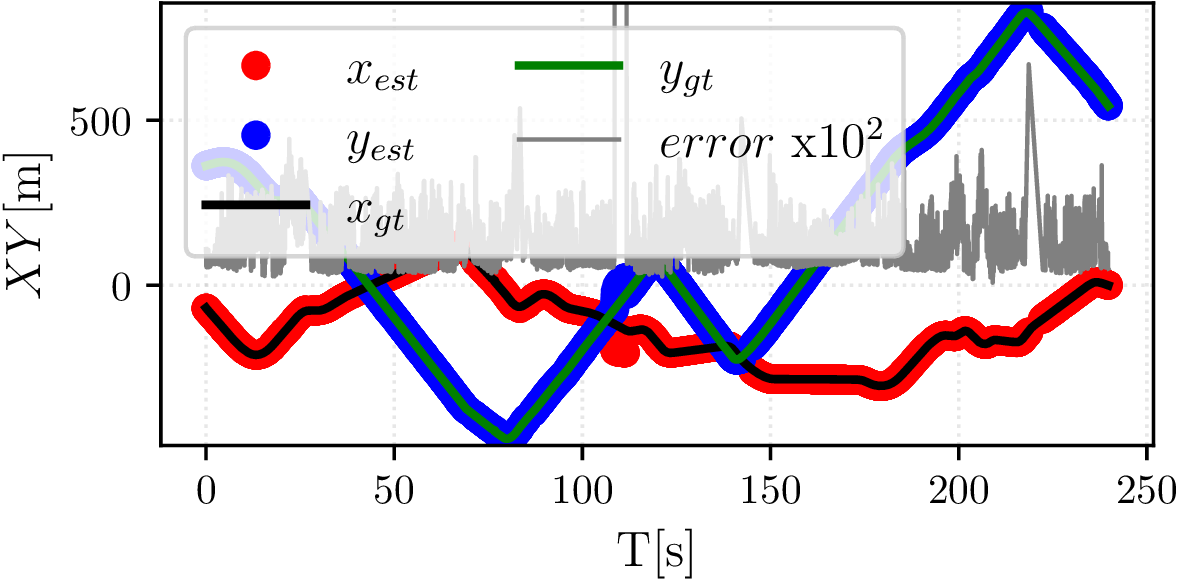} 
        % \caption{Estimated Vs ground-truth Position at LVRC}
        \caption{(c) LVRC}
        \label{fig:est_gt_pos_lvrc}
    \end{subfigure}
    \hfill 
    % Second subfigure
    \begin{subfigure}[b]{0.48\textwidth} 
        \centering
        \includegraphics[scale=0.6]{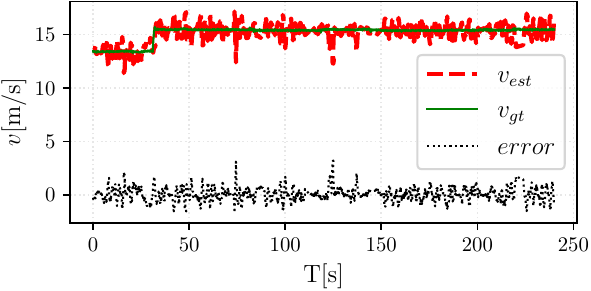}
        % \caption{Estimated Vs ground-truth Velocity at LVRC}
        \caption{(d) LVRC}
        \label{fig:est_gt_vel_lvrc}
    \end{subfigure}
    \vspace{-4pt}
    \caption{
    Comparison of estimated and ground-truth opponent states: (a) and (c) depict the UKF-estimated XY position ($x_{est}$, $y_{est}$) alongside the ground truth ($x_{gt}$, $y_{gt}$) from shared opponent localization, where the position error is defined as the Euclidean distance between them and scaled by a factor of 100 for clarity; (b) and (d) present the estimated velocity ($v_{est}$) versus the ground truth ($v_{gt}$), along with the absolute velocity error.
    % ; where error is $v_{gt} - v_{est}$.
    } 
    \label{fig:est_gt_pos_vel}
    \vspace{-0.15in}
\end{figure}

\begin{figure}[t!]
\centering
 \includegraphics[width=\linewidth]{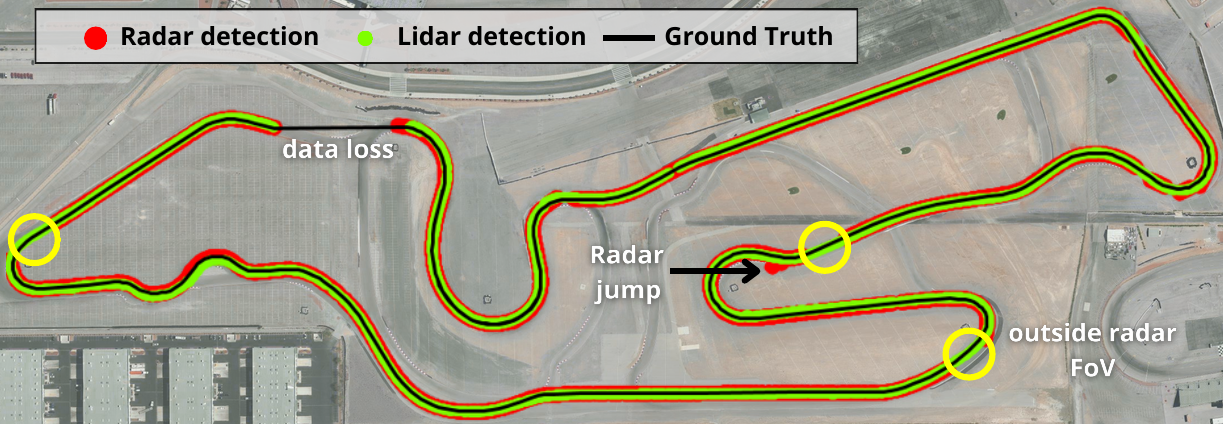}%
 % \includegraphics[width=\linewidth]{Racing_software_stack/Figs/detection/detection_with_bgc-1.png}%
% \vspace{-0.15in}
\caption{\small Opponent Tracking: radar-based tracking during the LVRC testing (running on our vehicle), while LiDAR-based tracking was tested on the recorded bag files.}
\label{fig:detection_track}
\vspace{-0.17in}
\end{figure}

In \texttt{ARS1}, we collected initial data involving an opponent vehicle.
% at KS. % we didn't mention ks above
Using this data, the radar-based opponent detection and tracking algorithm was evaluated offline to verify its functionality and real-time performance. Subsequently, in \texttt{ARS2}, this algorithm was tested on-track at LVMS during practice sessions involving two autonomous vehicles, monitored by the IAC chase car, where the opponent vehicle maintained varying speeds and relative distances from the ego vehicle. 
The algorithm operated reliably at a frequency of \SI{20}{\Hz}, consistently detecting opponents at distances of up to \SI{200}{\meter} across a range of speeds between \SI{7}{\meter\per\second} and \SI{28}{\meter\per\second}.
The \texttt{ARS2} detection algorithm demonstrated robust performance under dynamic conditions, with an opponent velocity error of $-0.007 \pm 1.35\,\si{\meter\per\second}$ and a distance error of $0.82 \pm 0.91\,\si{\meter}$ (see Fig.~\ref{fig:est_gt_pos_lvms} and Fig.~\ref{fig:est_gt_vel_lvms}). 
% The radar-based opponent detection was further improved. This improvement involved two key modifications: first, incorporating logic to detect when the opponent car is stopping based on its velocity and acceleration; and second, applying a moving average filter to smooth the opponent's estimated velocity.
The radar-based opponent detection was further improved through two key modifications: (i) the addition of a logic to detect when the opponent car is stopping based on its velocity and acceleration, and (ii) the application of a moving average filter to smooth the opponent's estimated velocity.
This enhanced estimation was validated during testing at LVRC (Figs.~\ref{fig:est_gt_pos_lvrc} and~\ref{fig:est_gt_vel_lvrc}), resulting in reduced errors in velocity ($0.05 \pm 0.53$~\si{\meter\per\second}) and mean distance error ($0.67 \pm 1.76\,\si{\meter}$) compared to earlier results at LVMS. While the standard deviation of velocity error showed particular improvement, the distance standard deviation was larger due to one big jump that occurred before the hairpin (see Fig.~\ref{fig:detection_track}).
% particularly in the standard deviation of velocity error, 
% however, distance standard deviation is larger due to the one jumb occured before the shicane (see Fig.~\ref{fig:detection_track}).
However, gaps in detection were observed at LVRC due to the radar’s limited field of view (FoV). 
% These gaps occur when the ego-car or opponent enters/exits a sharp curve, leading to the opponent car lying outside the radar FOV, see highlighted yellow circles in Fig. \ref{fig:detection_track}. 
These gaps occurred when either the ego vehicle or the opponent entered or exited a sharp curve, causing the opponent to fall outside the radar FoV (see highlighted yellow circles in Fig.~\ref{fig:detection_track}).
% To mitigate this issue, LiDAR tracking was implemented as an alternative approach. However, real-time execution on the vehicle's PC with all other modules running resulted in latencies exceeding 1 second. Offline validation using recorded bag data demonstrated an average pose error of $1.89 \pm 1.59,\mathrm{m/s}$, and mean velocity error of $0.49 \pm 1.50,\mathrm{m/s}$, see Fig.~\ref{fig:detection_track}

% \youwei{per review: The LiDARs have an update rate of 10 Hz --- whereas the
% radars are at 20 Hz. Will the relatively slower update
% frequency of LiDARs affect the performance for the
% LiDAR-radar fusion opponent estimation (in ARS3)}

% % lidar metrics
% % Average pose error: 1.89 ± 1.59 m
% % Average velocity error: -0.49 ± 1.50 m/s

\captionsetup[subfloat]{labelformat=empty}

% \begin{figure}[t!]
% \centering
%  \includegraphics[scale=0.85]{Racing_software_stack/Figs/detection/trajectory.pdf}%
% % \vspace{-0.32in}
% \caption{\small Opponent Tracking: radar-based tracking during the LVRC testing (running on car), while lidar-based tracking tested on the recorded bag file.}
% \label{fig:detection_track}
% \vspace{-0.25in}
% \end{figure}

% \begin{figure}[h!]
%     \centering
%     % First subfigure
%     \begin{subfigure}[b]{0.48\textwidth}
%         \centering
%         \includegraphics[scale=0.6]{Racing_software_stack/Figs/detection/distance_and_velocity.pdf} 
%         \caption{distance-velocity}
%         \label{fig:acc1}
%     \end{subfigure}
%     \hfill 
%     % Second subfigure
%     \begin{subfigure}[b]{0.48\textwidth} 
%         \centering
%         \includegraphics[scale=0.6]{Racing_software_stack/Figs/detection/velocity.pdf}
%         \caption{throttle-brake}
%         \label{fig:acc2}
%     \end{subfigure}
%     \caption{Acc.} 
%     \label{fig:acc}
% \end{figure}

% \noindent \bfit{Planning and Control.}
\subsection{Planning and Control}
\begin{figure}[htbp] % [htbp] is optional, for figure placement
    \centering % Center the entire figure environment

    \begin{subfigure}{0.48\linewidth} % Adjust width as needed, leave some space between
        \centering
        \includegraphics[width=\linewidth]{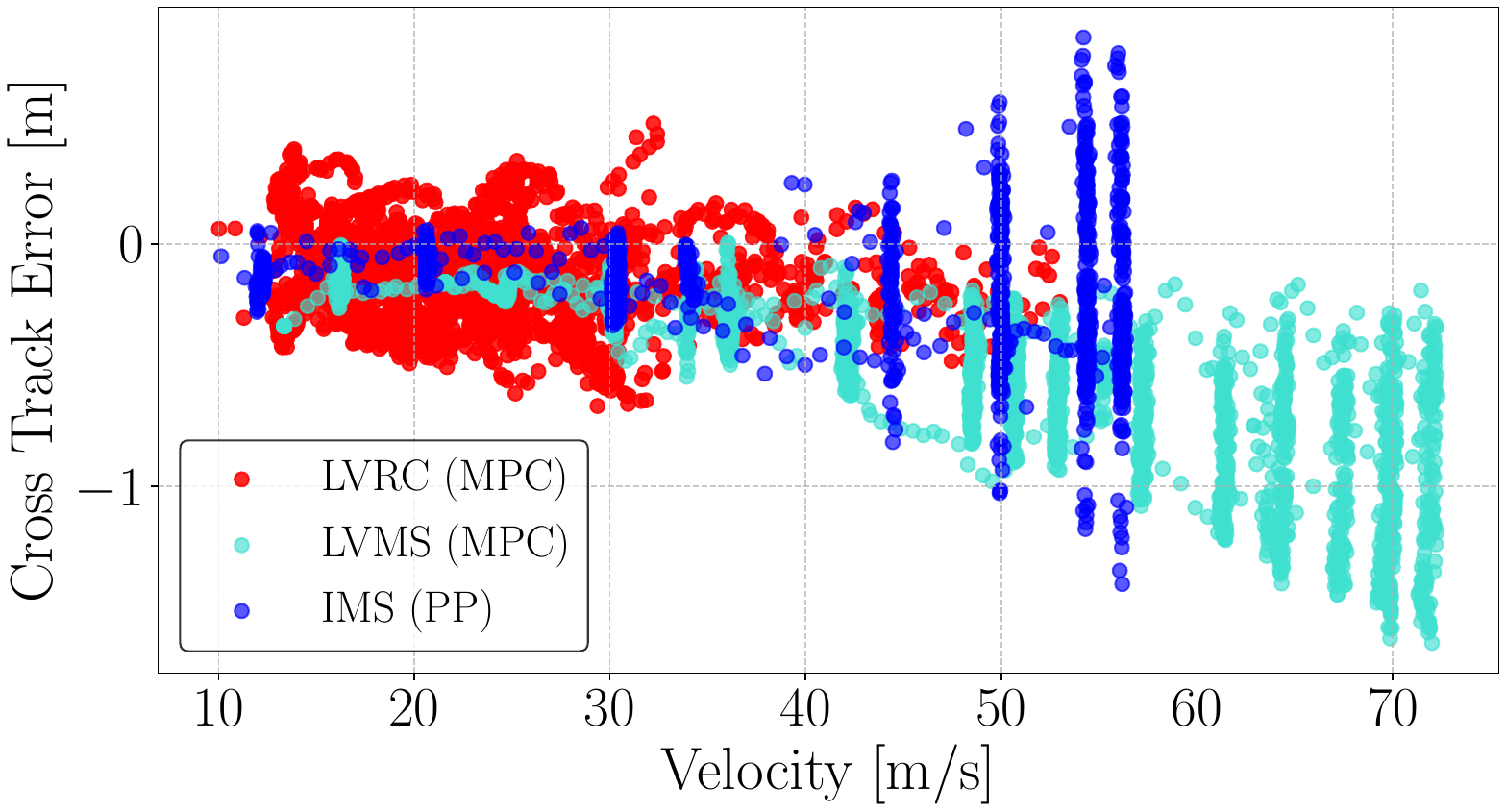}
        \caption{(a) Cross-Track Errors.}
        \label{fig:cte-vs-velocity-plot}
    \end{subfigure}
    \hfill % This command adds horizontal space between the subfigures
    \begin{subfigure}{0.48\linewidth} % Adjust width as needed
        \centering
        \includegraphics[width=\linewidth]{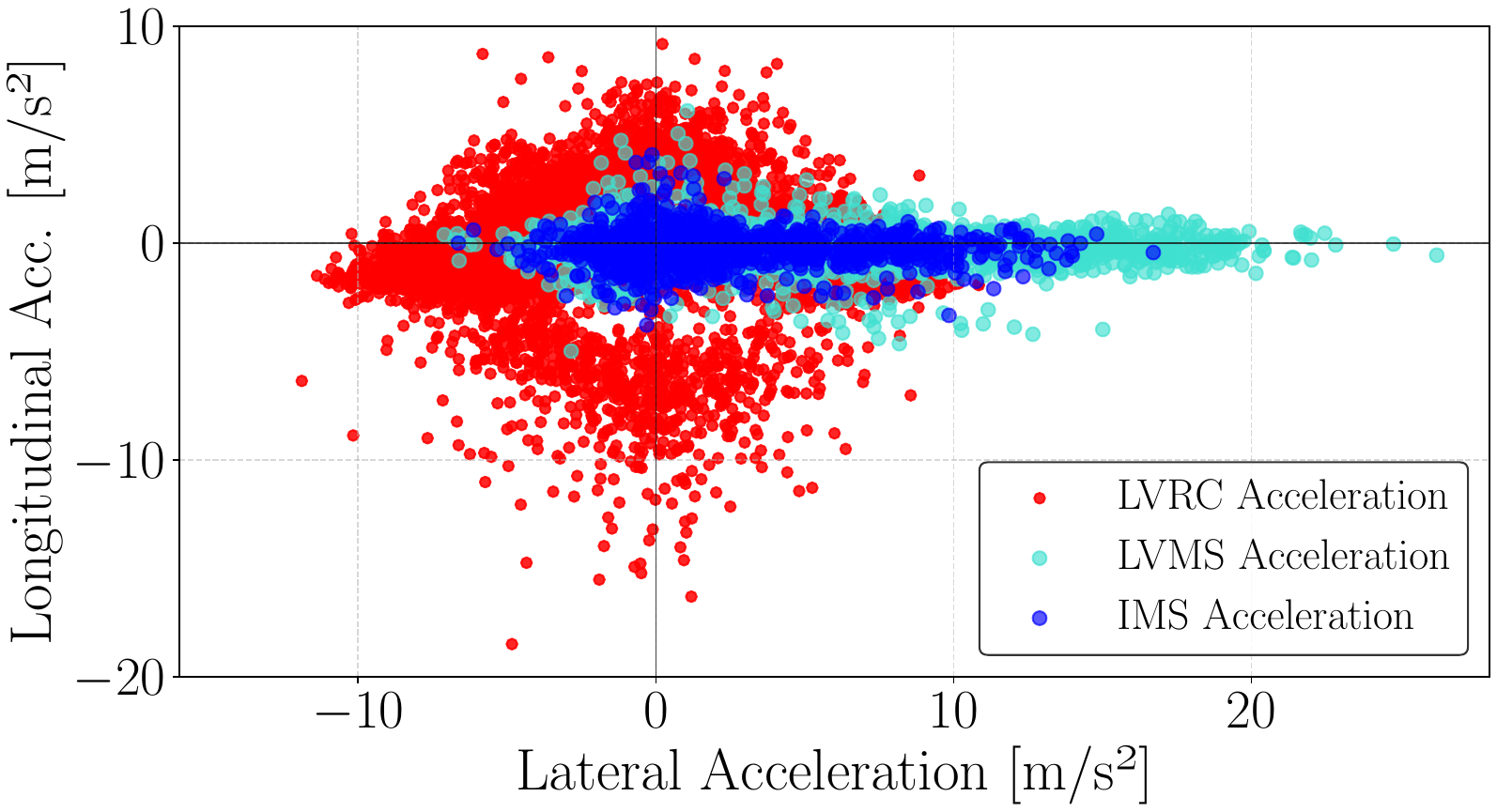}
        \caption{(b) GG Diagram.}
        \label{fig:gg_diagram} % Changed label to be more specific
    \end{subfigure}
    \vspace{-4pt}
    \caption{MPC and PP controllers performance on oval (IMS, LVMS) and road-course (LVRC) tracks: (a) trajectory tracking vs velocity and (b) lateral–longitudinal acceleration (GG diagram).} % Overall caption for the figure
    \label{fig:combined_plots} % Overall label for the figure
    \vspace{-0.15in}
\end{figure}

In \texttt{ARS1}, our autonomous racing vehicle achieved a maximum speed of \SI{205}{\kilo\meter\per\hour} while using Pure Pursuit (PP) for steering control. The cross-track error (i.e., lateral deviation from the pre-planned raceline) ranged from \SI{-1.5}{\meter} to \SI{0.6}{\meter}, while the heading error (i.e., deviation from the raceline heading at the closest point) varied between \SI{-2.0}{\degree} and \SI{2.25}{\degree}. While the vehicle remained stable throughout operation, oscillations were observed. 
% The maximum lateral acceleration reached \SI{18}{\meter\per\second\squared}, a critical factor for vehicle dynamics and tire limits.
The maximum lateral acceleration, a critical factor for vehicle dynamics and tire limits, reached \SI{18}{\meter\per\second\squared}.
In contrast, longitudinal acceleration was negligible on ovals due to constant high speeds.
In \texttt{ARS2}, lateral control was enhanced using an MPC based on the single-track dynamic model with banking compensation~\cite{rajamani2011vehicle}. This achieved a maximum speed of \SI{260}{\kilo\meter\per\hour}, with a cross-track error between \SI{-1.6}{\meter} and \SI{-0.4}{\meter} and a heading error within \SI{\pm 1}{\degree}. Maximum lateral acceleration increased to \SI{28}{\meter\per\second\squared}.
At LVRC, a road-course track, the vehicle's behavior differed significantly from that on ovals. Specifically, cross-track error ranged from \SI{-0.6}{\meter} to \SI{0.5}{\meter}, and heading error varied within \SI{\pm 4}{\degree} at a maximum speed of \SI{191}{\kilo\meter\per\hour}. The performance of both controllers, in terms of cross-track error versus velocity during high-speed runs on all three tracks, is depicted in Fig.~\ref{fig:cte-vs-velocity-plot}. 
% Moreover, road-course driving required alternating lateral acceleration (peak: \SI{10}{\meter\per\second\squared}) and substantial longitudinal braking (peak deceleration: \SI{-15}{\meter\per\second\squared}). 
Racing on the road course inherently involves dynamic changes in acceleration. The vehicle experienced varying lateral accelerations from alternating turns, reaching a peak magnitude of \SI{10}{\meter\per\second\squared}. Longitudinally, the vehicle consistently switched between accelerating and braking, which resulted in a peak deceleration of \SI{-15}{\meter\per\second\squared}.
The differences in acceleration behavior between ovals and road-course tracks can be seen in Fig.~\ref{fig:gg_diagram}.

% \begin{figure}[htbp] % [htbp] is optional, for figure placement
%     \centering % Center the entire figure environment

%     \begin{subfigure}{\linewidth} % Subfigure takes the full available width
%         \centering
%         \includegraphics[width=0.95\linewidth]{Racing_software_stack/Figs/opp_to_dist_d.pdf} % Adjust width as needed
%         % \caption{Caption for the first figure (e.g., Opponent to Distance)}
%         \label{fig:opp_to_dist}
%     \end{subfigure}

%     \vspace{-0.08cm} % Adjust this value to control the vertical space (e.g., 1cm, 10pt, 0.5in)

%     \begin{subfigure}{\linewidth} % Subfigure takes the full available width
%         \centering
%         \includegraphics[width=0.95\linewidth]{Racing_software_stack/Figs/throttle_brake_acc_d.pdf} % Adjust width as needed
%         % \caption{Caption for the second figure (e.g., Throttle, Brake, and Acceleration)}
%         \label{fig:throttle_brake_acc}
%     \end{subfigure}

%     \caption{Overall caption for both figures stacked together.} % Overall caption
%     \label{fig:stacked_vehicle_data} % Overall label
% \end{figure}

% \vspace{-0.1in}
\begin{figure}
    \centering
    \includegraphics[scale=0.4]{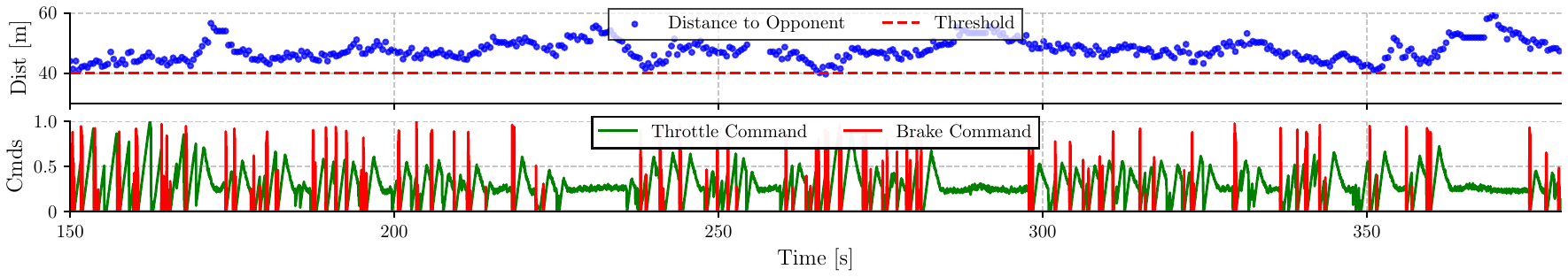}
    \vspace*{-2pt}
    \caption{ACC performance showing curvilinear distance to the opponent (top) and the resulting normalized control commands (bottom).}
    \vspace{-13pt}
    \label{fig:acc-performance}
\end{figure}

% Our ACC was validated during LVRC testing, with Fig. \ref{fig:acc-performance} demonstrating braking activation when curvilinear distance to leading vehicles approached safety thresholds. The car was first commanded a constant velocity of \SI{60}{\kilo\meter}, however, the ACC was commanding a lesser velocity and the controller was reacting to these commands. 

The ACC was validated during testing at LVRC. Its effective operation is demonstrated in Fig.~\ref{fig:acc-performance}, which illustrates brake activation as the curvilinear distance to a leading vehicle decreased to the predefined safety thresholds. In the depicted scenario, the vehicle was initially commanded to maintain a constant velocity of \SI{60}{\kilo\meter\per\hour}. Subsequently, as the lead vehicle was detected and the gap reduced, the ACC system appropriately commanded a lower velocity. As a result, the longitudinal controller successfully tracked this updated setpoint from the ACC, actuating the brakes to ensure a safe following distance.
% , thus validating the system's designed functionality.

\begin{figure}[htbp]
    \centering
    \includegraphics[width=0.95\linewidth]{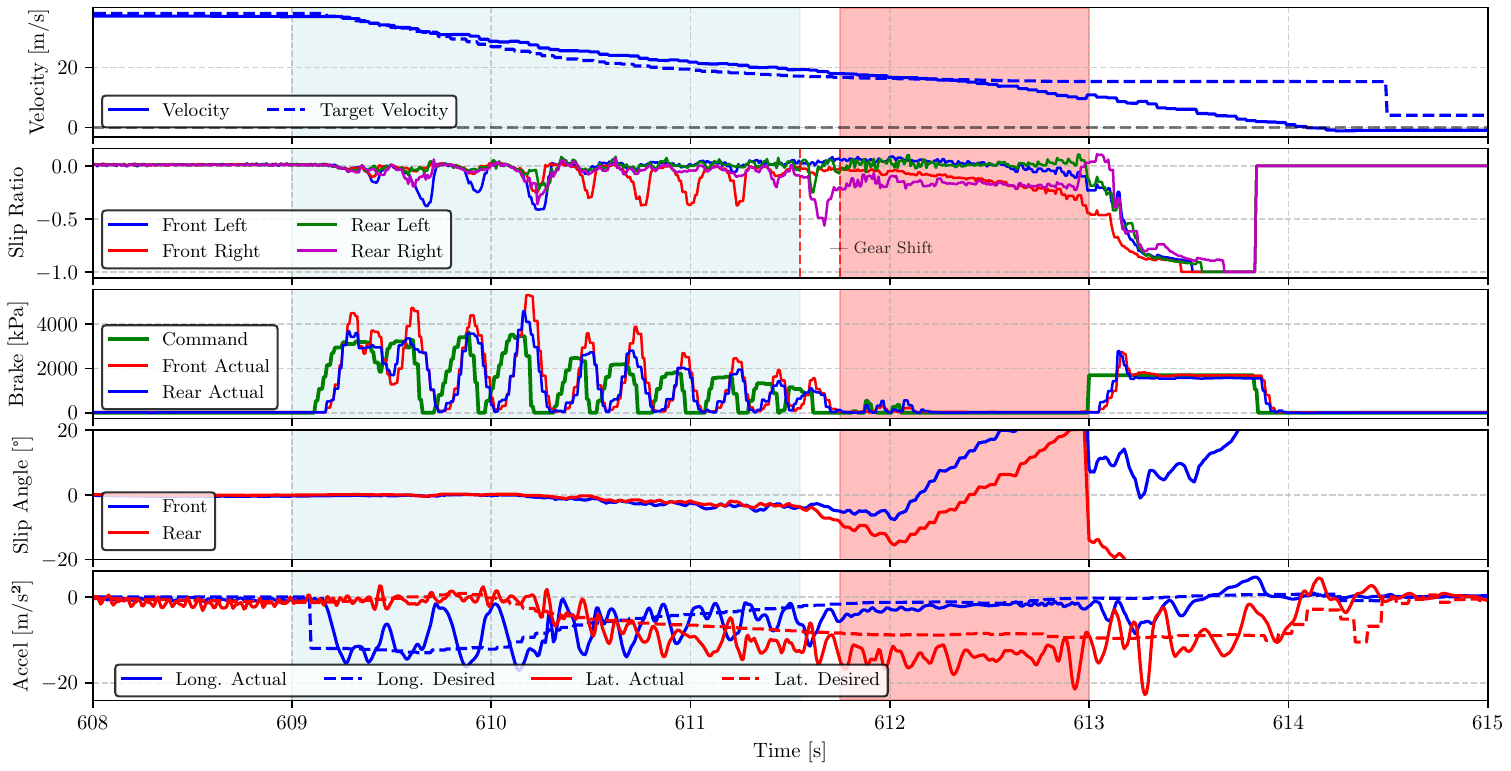}
    \caption{\small Vehicle data during an ABS-induced spin event. The plots display: 
    Vehicle and target velocities, tire slip ratios, commanded versus actual brake pressures, estimated slip angles~\cite{rajamani2011vehicle}, and 
    target and IMU-measured actual accelerations. 
    Blue shaded regions indicate ABS activation, while red shaded regions denote the vehicle spin. An increase in rear axle slip ratio is observed during gear shifting. Lastly, a safety check activated the brakes.}
    \label{fig:abs-spin}
    \vspace{-10pt}
\end{figure}

A critical instability event was observed during ABS evaluation on our final practice day at LVRC. The sequence initiated when the vehicle commenced deceleration for an upcoming turn. However, due to suboptimal tuning, the ABS failed to deliver the commanded deceleration, causing the vehicle to approach the turn entry at a velocity significantly higher than intended.
This excessive entry speed led to pronounced understeer. In this condition, the front tires exceeded their capacity for lateral grip before the rear tires, resulting in the vehicle's trajectory becoming wider than that dictated by the steering input; essentially, the vehicle deviates outward from the desired cornering arc. Simultaneously, slip angles, defined as the angle between the direction a tire is pointing and its actual direction of travel, began to increase, particularly at the front axle. It is crucial to distinguish slip angles, which are primarily associated with \textit{lateral force} generation during cornering, from slip ratios, which are associated with \textit{longitudinal forces}. While the initial failure involved ABS performance related to slip ratios (blue area in Fig.~\ref{fig:abs-spin}), the subsequent loss of control was characterized by excessive slip angles (red area in Fig.~\ref{fig:abs-spin}). 
Concurrently, two factors exacerbated the situation. First, as the vehicle's speed decreased, the gear controller, unaware of the high slip angles and the demanding high-curvature maneuver, commanded a downshift, causing high slip ratios at the rear tires. Second, the lateral MPC, attempting to correct the increasing cross-track error caused by understeer, commanded a progressively larger steering angle. The combination of these events led to a rapid increase in tire slip angles, indicating the onset of a spin. Ultimately, the high steering command, coupled with the vehicle's compromised state, resulted in a complete spin. A video of this incident is provided in the supplementary material for further analysis.

\subsection{Preliminary Testing Results of \texttt{ARS3}} 
Different from \texttt{ARS2} which was fully validated on the vehicle, certain new components of \texttt{ARS3}'s such as the LiDAR methods were validated only using offline data at the current stage, due to the limited access to tracks and vehicles in this racing season.
%In real-world autonomous racing, testing and tuning are inherently constrained by limited access to tracks and vehicles. As a result, \texttt{ARS2} was validated directly on the vehicle whereas \texttt{ARS3}'s LiDAR's methods were validated using offline data so far. 
% In contrast, due to time and computational constraints for \texttt{ARS3}, 
For \textit{state estimation}, LiDAR odometry computation time recorded $0.1 \pm 0.06~\si{\second}$ per frame on a Intel i9-14900KF CPU with four threads allocated. 
% For \textit{state estimation}, 
The results show better yaw angle accuracy, contributing to improved safety. Previously, GPS-based estimation caused up to three yaw jumps per lap, each reaching approximately \SI{2}{\degree}. With LiDAR, the yaw angles remain consistent throughout the lap, and the resulting map shows visually stable track boundaries, indicating reliable pose estimation.
For \textit{perception}, the LiDAR-based opponent tracking system was validated on recorded bag files, demonstrating a mean pose error of $1.89 \pm 1.59~\si{\meter\per\second}$ and a mean velocity error of $0.49 \pm 1.50~\si{\meter\per\second}$.
% (see Fig.~\ref{fig:detection_track}). 
However, real-time deployment on the vehicle’s onboard computer, while running all modules concurrently, introduced latency exceeding 1 second.
% Current work focuses on resolving these computational constraints to enable comprehensive testing of the integrated radar-LiDAR tracking framework.
% lidar metrics
% Average pose error: 1.89 ± 1.59 m
% Average velocity error: -0.49 ± 1.50 m/s
For \textit{control}, the MPPI controller was tested on track using a kinematic model at LVMS, achieving a speed of \SI{72}{\kilo\meter\per\hour}, and similarly tested at LVRC with comparable speeds. However, due to limited time for tuning, the performance was not sufficient to safely increase the speed further. Improving the performance of this new controller is our next goal.  
% \hassan{Any predictin experiments on the data or simulation?}.
% It is worth to note that our team evaluated several foundation model-based methods for lane detection; however, with inference speeds averaging around 2 seconds per image on an RTX 4080, these approaches proved impractical given the AV24's hardware constraints.
% Lastly, in \texttt{ARS3}, a \textit{Frenet-Frame} local planner is used for overtaking \cite{werling2010optimal}, while utilizing MPPI with the full single-track dynamic model as the main controller followed by a feedforward-feedback longitudinal controller that leverages the engine map to convert acceleration commands into throttle or brake inputs. These methods are currently integrated into our stack and tested in a Unity-based simulator \cite{autonomalabsOverviewAWSIM}. 
% It is worth to note that 

% Our team plans to validate the latest stack version, \texttt{ARS3}, at LVRC in April 2025. While all modules are implemented, further tuning based on data analysis and simulation is needed before track testing. On-track experiments will feature specialized maneuvers to enhance vehicle dynamics modeling, fine-tune control parameters, and execute high-speed overtakes. We also plan to re-run our previous \texttt{ARS} versions at LVRC and present both qualitative and quantitative comparisons of all three versions on the same road-course track. We also plan to share a dataset from the three tracks after LVRC testing. 
% \textcolor{red}{Should any testing efforts fail, we will include a comprehensive comparison of the methods employed.}

\section{Experimental Insights}\label{sec:5}
% In this section, insights and discussion from our experiments are presented.
We also present key insights and discussions based on our experimental results.
% and detail our future work towards \texttt{ARS3} full implementation. 
% \youwei{The intro should change.}

% \noindent \bfit{Summary of Differences Between \texttt{ARS} Iterations}
% % \mahmoud{we can use a table to give simple overview of our journey}
% \vspace{-0.2in}
% \begin{table}[htbp]
% \centering
% {\small
% \caption{Summary of Differences Between ARS Iterations}
% \label{tab:ars_summary}
% \rowcolors{1}{gray!25}{white}
% \begin{tabular}{|l|c|c|c|}
% \hline
% \rowcolor{gray!50}
% \textbf{Feature} & \bfit{ARS1} & \bfit{ARS2} & \bfit{ARS3} \\ \hline
% \textbf{Racing Scenario} & SV & SV and 
%  H2H-trial & H2H \\ \hline
% \textbf{Track} & IMS & LVMS & LVRC \\ \hline
% \textbf{Track Hours} & 11 hours & 6.5 hours & Approx.6 hour \\ \hline
% \textbf{Kilometers Traveled} & 325 km & 166 km & TBD \\ \hline
% \textbf{Top Speed} & 205 km/h & 260 km/h & TBD \\ \hline
% \textbf{Estimation} & GPS/INS & GPS/INS & GPS/INS \& Lidar \\ \hline
% \textbf{Perception} & N/A & Radar-based & Radar-Lidar fusion \\ \hline
% \textbf{Prediction} & N/A & UKF & MPC-based \\ \hline
% \textbf{Controller} & PP & PP and MPC & PP, MPC, and MPPI \\ \hline
% % \textbf{Validation Status} & Validated & Validated & To be validated \\ \hline
% \end{tabular} 
% % \vspace{-12pt}
% }
% \end{table}

% \noindent \bfit{Sensor Redundancy in State Estimation.}
\subsection{State Estimation with Redundant Sensors}
% As introduced in Section~\ref{sec:technical_approach}, each sensor has its backups, but the inconsistent extrinsic calibrations pose a significant challenge for sensor fusion. Common practices rely on supplementary sensor motion measurements to enhance motion estimation accuracy; however, this approach restricts the pose transition between sensors at high vehicle speeds when the primary sensor fails (e.g., due to hardware connection loss). In terms of transition between different GNSS devices, we mainly relied on IMQ smoothing to avoid sudden jumps. We will investigate a consistent transition algorithm in high-speed situations. Moreover, exteroceptive sensor (e.g., LiDAR, camera) odometry~\cite{ORBSLAM3_TRO, vizzo2023ral} typically suffers from scale ambiguities, further complicating sensor transitions. While coupling sensors can improve accuracy under normal conditions, we will investigate methods for ensuring safe and smooth transitions in \texttt{ARS3}. Safety indicates the solution ambiguity, when the loss function converges but the solution is unrealistic. We plan to investigate cross-validation techniques, including (1) versatile feature selection online~\cite{fontanRSS2024} and (2) robust map tracking with different sensors.
As introduced in Section~\ref{sec:technical_approach}, each sensor has its backups, but inconsistent extrinsic calibrations pose a significant challenge for sensor fusion. Common practices rely on supplementary sensor motion measurements to enhance motion estimation accuracy; however, this approach restricts pose transitions between sensors at high vehicle speeds when the primary sensor fails (e.g., due to hardware connection loss). Regarding transitions between different GNSS devices, we mainly rely on IMQ smoothing to avoid sudden jumps. We will investigate a consistent transition algorithm for high-speed situations. Moreover, exteroceptive sensor (e.g., LiDAR, camera) odometry~\cite{ORBSLAM3_TRO, vizzo2023ral} typically suffers from scale ambiguities, further complicating sensor transitions. While coupling sensors can improve accuracy under normal conditions, we will investigate methods to ensure safe and smooth transitions in \texttt{ARS3}. Here, safety refers to solution ambiguity, i.e., when the loss function converges but the solution is unrealistic. 
% We plan to explore cross-validation techniques, including (i) versatile feature selection online~\cite{fontanRSS2024} and (ii) robust map tracking with different sensors.
We plan to explore cross-validation techniques, including versatile feature selection online
% ~\cite{fontanRSS2024}
and map tracking with different sensors.

% \noindent \bfit{Perception Hallucinations.}
\subsection{Perception Hallucinations}
False-positive detections typically arise from environmental factors such as multipath radar signals or sensor noise, potentially causing unintended behavior within the planning module. In \texttt{ARS2}, track boundaries and historical detection data were utilized to mitigate these false detections. 
% Unlike radar, lidar is not susceptible to multipath effects, suggesting that lidar-radar tracking will substantially reduce false positives and enhance perception robustness. Fig. \ref{fig:detection_track} demonstrates this improvement where lidar-based tracking compensates for radar FoV gaps (yellow circles) and eliminates tracking-jumps in radar-based tracking caused by multipath interference. 
Unlike radar, LiDAR is not susceptible to multipath effects, suggesting that LiDAR-radar fusion could significantly reduce false positives and improve perception reliability. As demonstrated in Fig.~\ref{fig:detection_track}, the LiDAR-based tracking component effectively compensates for radar FoV limitations (highlighted in yellow circles) while eliminating radar tracking discontinuities caused by radar multipath interference.
The current LiDAR-based tracking system could not be deployed in real-time due to computational constraints—execution times exceeded \SI{1}{\second}.
% when running concurrently with other modules. 
Future work will focus on optimizing computational efficiency to enable real-time fused LiDAR-radar tracking.

% \noindent \bfit{Dynamic and Uncertainty-Aware Control Framework.}
\subsection{Planning-Control Framework}
% \youwei{Do we need bold texts here? That is not consistent with previous sections\\}
In \texttt{ARS2}, the planning architecture integrates an \textit{offline-generated optimal raceline} with an \textit{online module}. 
The online component encompasses a \textit{behavioral planner} for multi-vehicle interactions, tailoring its decisions based on competition rules, and an ACC system. For high-speed multi-vehicle racing scenarios on road-courses, a \textit{local motion planner} is needed within this online framework to generate dynamically feasible trajectories for overtaking maneuvers as the assumption that both cars travel on parallel racelines is not always valid. The control system employs a decoupled approach. \textit{Longitudinal control} is achieved using two PID controllers, one for throttle and another for braking, complemented by an ABS designed to maximize deceleration. For \textit{lateral control}, a linear MPC is implemented.
While this decoupled strategy has proven effective in many scenarios, it has inherent disadvantages that can, under certain conditions, lead to vehicle instability, as demonstrated by the spin incident shown in Fig.~\ref{fig:abs-spin} at LVRC. Our future efforts therefore include tuning the coupled MPPI controller that considers nonlinear vehicle dynamics.
%, were promising but limited by restricted track access and the considerable time required for comprehensive tuning.

% \noindent \bfit{Performance Improvement against Traversed Miles.}
% \subsection{Towards \texttt{ARS3} full implementation}
\subsection{Conclusion and Future Work}
This paper has presented performance results for the \texttt{ARS1} and \texttt{ARS2} systems on both oval tracks and road courses. The development and comprehensive validation of our subsequent system, \texttt{ARS3}, encountered practical limitations, primarily restricted track access and computational constraints. Future work will therefore be directed towards overcoming these challenges and enhancing \texttt{ARS3}'s capabilities. Key areas of focus will include addressing computational issues through the optimization of sensor drivers and further refining the vehicle dynamics model along with the MPPI controller parameters. Additionally, efforts are being directed towards improving the precision of opponent vehicle trajectory prediction and advancing the local planner's ability to generate overtaking maneuvers effectively.

% Our approach has demonstrated exceptional efficiency in IAC challenges, with the vehicle successfully logging \SI{490}{\kilo\metre} in approximately 17.5 hours—the shortest practice time among existing global teams to achieve a top speed of \SI{260}{\kilo\metre\per\hour}. % at LVMS. 
% (11 hours for \texttt{ARS1} and 6.5 hours for \texttt{ARS2}). 
% This includes time spent troubleshooting hardware, acquiring operational expertise, and moving in the pits at low speed. 
% Managing limited track time was crucial to achieve research and competition goals. Our strategy involved initially exploring multiple methods, then concentrating on the most promising approach, analyzing data and tracking errors, and evaluating performance before finalizing settings for race day.

% \noindent \bfit{Aggressiveness and Profits.}
% \\ \\ \\ \\ \\ \\ \\

\section{Acknowledgement}
{\small
We acknowledge support from NSF Grants \#2047169, \#2006886, and the IU Luddy Autonomous Racing Initiative for funding related robotics research and racing event travel. Paul D. Coen was supported by the NISE/PhD Fellowship at NSWC Crane during his participation in 2024.  We thank Shaekh Shithil and Paul O. Quigley for their assistance with racing experiments, and we are grateful to the technical and logistics personnel of the Indy Autonomous Challenge for their support.
}
% \newpage
% \newpage
{\small
\bibliographystyle{IEEEtran}
\bibliography{references}
}

\end{document}